\documentclass[runningheads]{llncs}

\usepackage{eccv}

\usepackage{eccvabbrv}

\usepackage{graphicx}

\usepackage{tikz}
\usepackage{comment}
\usepackage{color}
\usepackage{booktabs}
\usepackage{colortbl,array,xcolor}
\usepackage{url}
\usepackage{multirow}
\usepackage{here}
\usepackage{amssymb}
\usepackage{pifont}
\newcommand{\cmark}{\ding{51}}%
\newcommand{\xmark}{\ding{55}}%

\usepackage[whole]{bxcjkjatype}
\usepackage{bm}

\usepackage[accsupp]{axessibility}  

\usepackage{hyperref}

\usepackage{orcidlink}

\begin{document}
\def\ECCVSubNumber{12}

\title{ZoDi: Zero-Shot Domain Adaptation with Diffusion-Based Image Transfer} 

\titlerunning{ZoDi: Zero-Shot Domain Adaptation with Diffusion-based Image Transfer}

\author{Hiroki Azuma\inst{1, 2} \and
Yusuke Matsui\inst{1} \and 
Atsuto Maki\inst{2}
}

\authorrunning{H. Azuma et al.}

\institute{The University of Tokyo \and
KTH Royal Institute of Technology}

\maketitle

\begin{abstract}
Deep learning models achieve high accuracy in segmentation tasks among others, yet domain shift often degrades the models' performance, which can be critical in real-world scenarios where no target images are available. 
This paper proposes a zero-shot domain adaptation method based on diffusion models, called {\it ZoDi}, which is two-fold by the design: zero-shot image transfer and model adaptation. First, we utilize an off-the-shelf diffusion model to synthesize target-like images by transferring the domain of source images to the target domain. In this we specifically try to maintain the layout and content by utilising layout-to-image diffusion models with stochastic inversion. Secondly, we train the model using both source images and synthesized images with the original segmentation maps while maximizing the feature similarity of images from the two domains to learn domain-robust representations. 
Through experiments we show benefits of {\it ZoDi} in the task of image segmentation over state-of-the-art methods. 
It is also more applicable than existing CLIP-based methods because it assumes no specific backbone or models, and it enables to estimate the model's performance without target images by inspecting generated images. Our implementation will be publicly available at \url{https://github.com/azuma164/ZoDi}.

\end{abstract}

\section{Introduction}




Deep neural networks have greatly advanced the performance of various recognition tasks in computer vision~\cite{NIPS2015-RPN,ronneberger2015unet,he2018maskrcnn,kirillov2019panoptic}. 
The recognition 
models typically perform well when the data distributions are consistent between the domains of training and testing set. 
Those come, however, with a downside that the performance 
may significantly drop when tested on out-of-distribution data~\cite{liang2024objectdetectorsopenenvironment,yang2024generalizedoutofdistributiondetectionsurvey,ovadia2019distributionshift,torralba2011bias,tommasi2015deeperbias}, which can be critical in real-world 
applications. 

Faced with this issue of domain shift, 
some work introduced domain adaptation techniques~\cite{long2015featurelevel,murez2017imagelevel,michieli2020gta5} 
trying to make good use of  
images from the target domain in an unsupervised way, 
i.e. by accessing
them without labels. 
There are various attempts for the challenge, such as feature-level~\cite{long2015featurelevel}, image-level~\cite{murez2017imagelevel}, and output-level adaptation~\cite{michieli2020gta5}. 
Recent research~\cite{luo2020adversarial,gong2022oneshot,wu2021style,benigmim2023datum} 
brought it to 
even stricter settings of one-shot unsupervised training where only one target image is 
made available. 
It is however not always possible to find a proper one.

\begin{figure*}[t]
\begin{center}
   \includegraphics[width=1.02\linewidth]{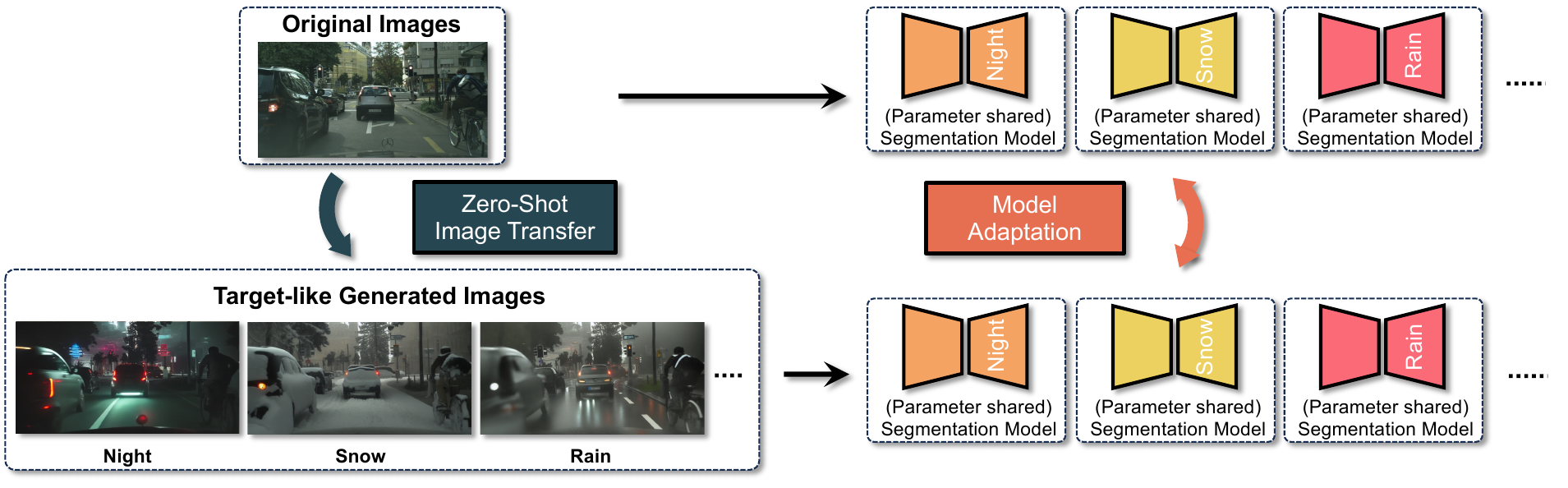}
   \caption{\textbf{The proposed training scheme -- ZoDi.} It comprises {\it zero-shot image transfer} and {\it model adaptation}; conducting image transfer to change the domain of original images to a target domain, and training a segmentation model using the transferred target-like images together with the original ones.
   See {\bf Fig.2} for a more detailed sketch.
    }
   \label{fig:abstract}
\end{center}
\end{figure*}

Zero-shot
domain adaptation in this respect is a promising 
direction for reducing the requirement in the case that real images from the target domain are unavailable. Some research~\cite{lengyel2021ciconv,luo2023similarity} introduced 
an effective approach to 
zero-shot {\tt day-night} adaptation 
although 
its usage is limited to that specific setting.  
Very recently, 
inspiring work in~\cite{vidit2023clipthegap,fahes2023poda} proposed utilizing the text features of prompts from CLIP~\cite{Radford2021-bi}, a vision-language pre-trained model,  
realising
more general zero-shot settings, 
and delivering decent results.
Nevertheless, 
they require freezing the image backbone of CLIP, which 
restricts the selection of the backbone and its performance. 
As they also 
adapt the model in the CLIP feature space, 
it may be problematic to estimate
the model's performance without target images.

In this paper, we are also concerned ourselves with zero-shot domain adaptation applicable to general settings, but without restrictions or uncertainty of results. 
Inspired by the strong zero-shot abilities of diffusion models, we propose a method called
\textbf{\underline{Z}}ero-sh\textbf{\underline{o}}t domain adaptation with \textbf{\underline{D}}\textbf{\underline{i}}ffusion-Based Image Transfer (ZoDi)
while focusing on the task of segmentation. 
As illustrated in Fig.~\ref{fig:abstract}, 
it comprises zero-shot image transfer and model adaptation.
Our basic strategy is to leverage powerful diffusion models~\cite{ho2020denoising,rombach2022stablediffusion} for zero-shot image transfer of source images to the target domain. 
That is, we first transfer images to the target domain using diffusion models although it involves a non-trivial question of how to keep the layout and content. We then train our model for segmentation while learning domain-robust representations by maximizing the similarity of image features from the two domains. 

In this framework, just as mentioned above,    
it remains as a challenge to maintain the image layout and content while changing the domain of the images.
For example, in existing diffusion-based image transfer methods~\cite{hertz2022prompttoprompt,mokady2022nulltext,Brooks2023instructpix2pix}, some objects in the original images often get eliminated, which can harm training segmentation models if directly applied. As a solution, we propose a novel image transfer method by adopting layout-to-image diffusion models~\cite{zhang2023adding,xue2023freestyle,li2023gligen,mo2023freecontrol} and stochastic inversion~\cite{zhang2023inversionbased}.
As for layout-to-image models, we employ pre-trained ControlNet~\cite{zhang2023adding} for its simplicity. To maintain the information of original images, we use real images as guidance and adapt stochastic inversion techniques presented in InST~\cite{zhang2023inversionbased}. 
Through experiments, we show that this image transfer method is capable of changing the image domain with little or no artifacts and thereby contributes to improving the model performance on segmentation tasks.

Since this method does not assume any specific backbone such as CLIP, ZoDi is available for existing daytime models. In addition, it is also possible to estimate the model's performance in advance by inspecting the generated images. 
Those are clear advantages over 
CLIP-based methods~\cite{vidit2023clipthegap,fahes2023poda}. To verify its effectiveness, we conducted various experiments on semantic segmentation in different settings: {\tt day}$\rightarrow${\tt night}, {\tt clear}$\rightarrow${\tt adverse weather}, and {\tt real}$\rightarrow${\tt game}. 
We observed 
consistent improvements over the models trained only by source images in all settings (e.g. +2.3 mIoU in {\tt day}$\rightarrow${\tt night} and +4.8 mIoU in {\tt clear}$\rightarrow${\tt snow}). 
To our surprise, it also outperforms DAFormer~\cite{hoyer2022daformer}, a strong baseline for unsupervised method, in some settings (+1.8 mIoU in {\tt clear}$\rightarrow${\tt snow} and +4.5 mIoU in {\tt clear}$\rightarrow${\tt rain}) 
without requiring 
any target images. 
These improved results in various settings are encouraging and demonstrate 
the benefits 
over existing methods. The contributions of the paper can be summarized as below:

\begin{itemize}
\item We introduce a zero-shot domain adaptation method based on diffusion model for the first time to our knowledge. We call it {\it ZoDi}. It consists of diffusion-based {\it image transfer} and similarity-based {\it model adaptation}.                                                                                         
\item
We propose to use layout-to-image diffusion models with stochastic inversion for high-quality image transfer, i.e. to maintain images' layout and content.
%
It helps train a segmentation model that allows state-of-the-art performance.

\item ZoDi does not assume any specific model backbone, unlike CLIP-based methods, which enables acting as a plug-and-play method to other existing daytime models. In addition, it enables estimating the model's performance without target images as we generate actual images.
\end{itemize}

\section{Related work}
\label{related}
\subsection{Zero-Shot Domain Adaptation}
Zero-shot domain adaptation assumes stricter settings with respect to  the conventional scenario; no images from the target domain are accessible. Given this, a relevant approach is to employ images in the target domain from unrelated tasks~\cite{peng2018zero,Wang_2019_ICCV,Wang_2020_ECCV} although it is not so evident when it comes to how to collect target images. 
Some research~\cite{lengyel2021ciconv,luo2023similarity} showed zero-shot {\tt day-night} adaptation assuming no target images but with specific limitation to the day-night settings. 
For a more general setting a few recent works~\cite{vidit2023clipthegap,fahes2023poda} exploit CLIP~\cite{Radford2021-bi}, a vision-language pre-trained model based on 400-M web-crawled image-text pairs. One of them~\cite{vidit2023clipthegap} uses the difference between the text features of source and target prompts to enable semantic augmentations. Another approach, P\O DA~\cite{fahes2023poda}, also utilizes CLIP for prompt-driven instance normalization to address zero-shot domain adaptation. Those methods however require freezing the image backbone of CLIP, which could restrict the selection of the backbone and its performance. It may not be also straightforward to estimate the model's performance without target images since they adapt the model in the CLIP feature space.

\subsection{Generative Models for Domain Adaptation}
To deal with the domain shift problems, some research~\cite{Wang_2019_ICCV,Wang_2020_ECCV,mohwald2023darkside} has used generative models to generate images similar to those in the target domain.
A common approach is to use GAN-based~\cite{goodfellow2014generative} generative models for the domain transfer. However, they are effective only with access to sufficient amount of images from the target domain for training  generative models.
Based on pre-trained diffusion models with huge capacity~\cite{ho2020denoising,rombach2022stablediffusion,Saharia2022-ma,ramesh2021dalle}, on the other hand, a few works ~\cite{benigmim2023datum,chopra2024sourcefree} explored to use them for domain adaptation. DATUM~\cite{benigmim2023datum}, one of the closest work to ours, utilizes pre-trained diffusion models for one-shot unsupervised domain adaptation. It fine-tunes text-to-image diffusion models with one image from the target domain using DreamBooth~\cite{Ruiz2022-oy}. It then generates target-like images and  apply other unsupervised methods to train the models. Despite the novelty, it should be noted that it needs a target image, requires heavy computation for the fine-tuning, and relies on other complicated unsupervised methods.


\subsection{Diffusion-based Image Style Transfer}
Some domain adaptation methods~\cite{Wang_2019_ICCV,Wang_2020_ECCV} including ours utilize image style transfer for changing the source images' domain to the target domain.
While image style transfer has been an important topics in computer vision, recent work~\cite{hertz2022prompttoprompt,mokady2022nulltext,zhang2023inversionbased} studied how to leverage the power of pre-trained diffusion models for the purpose. To be noted is that it is a non-trivial problem to keep the image's content in using diffusion models. Among a few attempts for this, Prompt-to-Prompt~\cite{hertz2022prompttoprompt} utilizes source attention maps to keep every pixel intact except the area to be changed. Mokady et al.~\cite{mokady2022nulltext} introduced null-text inversion which learns the unconditional textual embedding used for classifier-free guidance. The work in ~\cite{zhang2023inversionbased} proposed {\it stochastic inversion} which focuses on the impact of random noise in diffusion steps and adapts the initial input in the inversion-based style transfer method (InST). It is computationally efficient without requiring any optimization steps. Due to its simplicity and effectiveness, our method also adopts stochastic inversion.


\subsection{Layout-to-Image Diffusion Models}
In the literature there are some attempts~\cite{li2023gligen,zhang2023adding,xue2023freestyle,yang2023reco,chen2023geodiffusion,mo2023freecontrol} to employ diffusion models for generating images using text and some layouts, such as canny edge, depth maps, human pose, segmentation maps, or bounding boxes. 
While 
some work~\cite{yang2023reco,chen2023geodiffusion} uses layout-to-image diffusion models to generate high-quality object detection data, GeoDiffusion~\cite{chen2023geodiffusion} first showed that generated images by the bounding box layout improve detection models' performance. 
On the other hand, 
it has not been explored to use layout-to-image for segmentation tasks, whereas in this paper we utilize a strong layout-to-image models with segmentation maps. 
Using segmentation maps to generate images, we can use the original segmentation maps as the ground truth for synthesized images.

\vspace{2mm}
With respect to the related work, as a whole our method aims to enable model-agnostic general zero-shot domain adaptation by utilizing diffusion models for image transfer, where we transfer input images to any domain while maintaining their layout and content. This is then integrated into our model adaptation.

\section{Method}
\subsection{Preliminaries: Latent Diffusion Models}
Latent diffusion models~\cite{rombach2022stablediffusion, podell2023sdxl} perform the diffusion process in the latent space. First, it encodes images $x$ to the latent space $z_0 = E(x) \in\mathbb{R}^{h\times w\times c}$ and then obtains noisy samples by forward diffusion steps. As shown in~\cite{ho2020denoising,lu2022dpmsolver}, the forward diffusion has a closed-form solution to obtain the noisy sample $z_t\in\mathbb{R}^{h\times w\times c}$ in any time step $t \in\mathbb{N}$,
\begin{equation}
\label{eq:obtain_noise}
    z_t = \alpha_tz_0+\sigma_t\epsilon,
\end{equation}
where $\alpha_t, \sigma_t\in\mathbb{R}^{+}$ are determined by the noise schedules of the sampler. $\epsilon\in\mathbb{R}^{h\times w\times c}$ is random noise drawn from $\mathcal{N}(0, I)$.
After that, diffusion models decode the samples by repeating backward diffusion steps and produce images by decoding the latent vector. The objective function can be formulated as
\begin{equation}
    \mathcal{L}_\mathrm{LDM} = E_{z_0, \epsilon, t}\Vert\epsilon - \epsilon_\theta(z_t, t, c_p)\Vert^2,
\end{equation}
where $\epsilon_\theta$ is a denoising U-Net~\cite{ronneberger2015unet} model, $c_p\in\mathbb{R}^{d}$ is the embedding of the input text, and $\epsilon\sim\mathcal{N}(0, I)$.

\subsection{The Proposed Method: ZoDi}
\begin{figure*}[t]
\begin{center}
    \vspace{5mm}
   \includegraphics[width=0.99\linewidth]{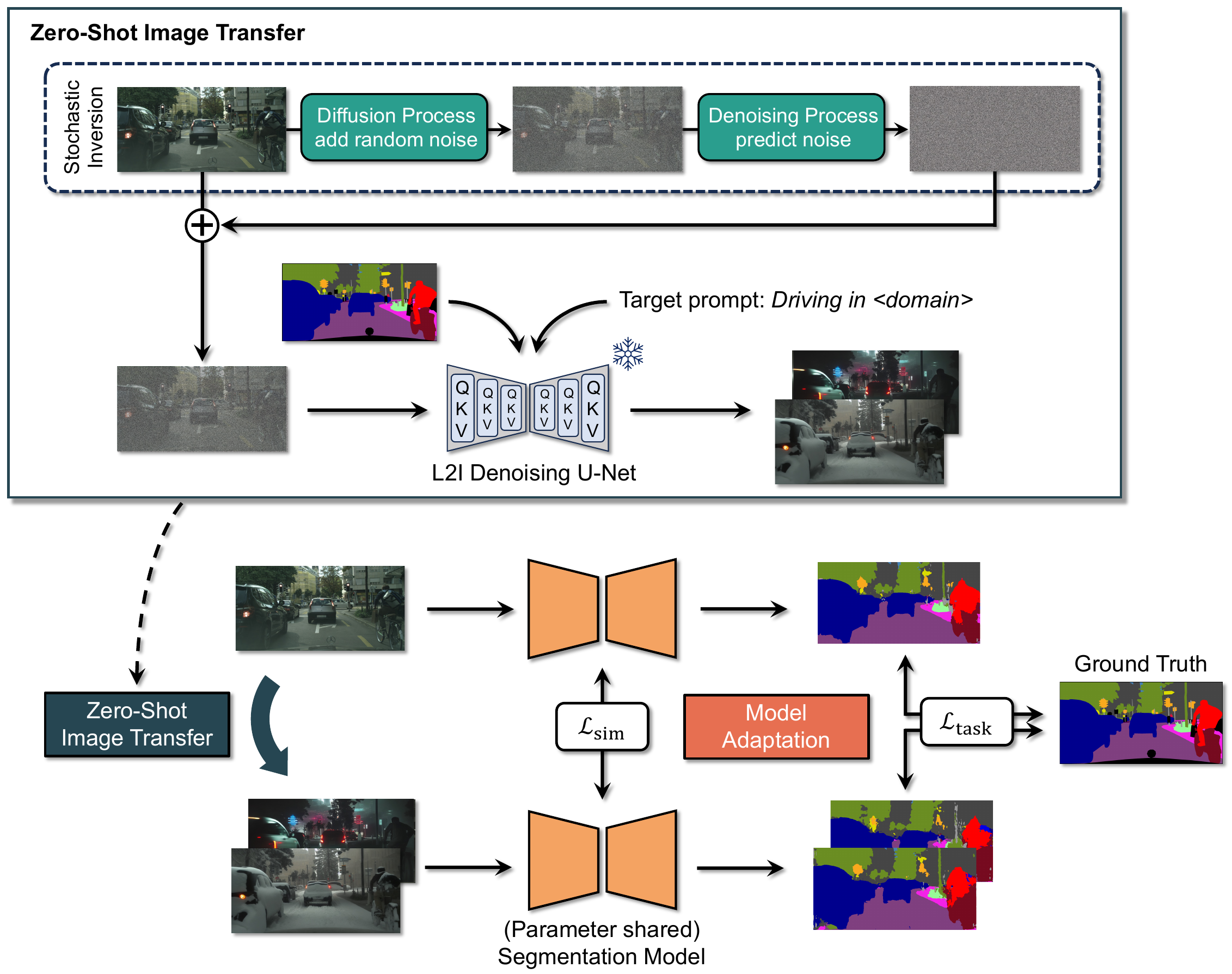}
   \vspace{5mm}
   \caption{\textbf{The architecture of ZoDi.} It consists of two components: zero-shot image transfer and model adaptation. First, for changing the domain of the original images, we design layout-to-image (L2I) diffusion models with stochastic inversion for zero-shot image transfer. We use the original segmentation maps and target prompt, e.g. ``driving in $<$domain$>$''. We then train the model with two losses: task loss and similarity loss.}
\label{fig:architecture}
\end{center}
\end{figure*}
As illustrated in Fig.~\ref{fig:architecture}, our method comprises two stages: zero-shot image transfer and model adaptation. 
That is, we use pre-trained diffusion models in order to obtain target-like generated images corresponding to source images by a zero-shot image transfer. We then train the segmentation models while maximizing the similarities between the features extracted from the original and generated images to make the feature space more robust against domain differences.

\subsubsection*{Zero-Shot Image Transfer:}

For zero-shot image transfer, 
it is crucial to maintain both the layout and content of the the images. 

To keep the layout, we employ {\it layout-to-image} (L2I) diffusion models. In those models, the predicted noise can be described as:
\begin{equation}
    \mathcal{L}_\mathrm{L2I} = E_{z_0, \epsilon, t}\Vert\epsilon - \epsilon_\theta(z_t, t, c_p, y)\Vert^2,
\end{equation}
where $y$ denotes the segmentation map for the input.
We adopt ControlNet~\cite{zhang2023adding} here for its simplicity\footnote{Other (L2I) diffusion models can be considered as replacement, e.g.  
GLIGEN~\cite{li2023gligen} or FreestyleNet~\cite{xue2023freestyle},
since our method is not specifically dependent on ControlNet.
}.

We also adopt the stochastic inversion proposed in InST~\cite{zhang2023inversionbased}, which helps maintain the content of the images, because we found that applying L2I models alone is not sufficient to prevent the generated images from collapsing on one way or another. Stochastic inversion focuses on the impact of random noise in diffusion steps. In this stochastic inversion, we aim to adapt the initial noise sample $z_k$ to $z_k'$. We first encode the original images 
and add noise for $k$-steps using Eq.~\ref{eq:obtain_noise} to obtain $z_k\in\mathbb{R}^{h\times w\times c}$.
Then we use the denoising U-Net $\epsilon_\theta$ to predict the noise $\epsilon_k\in\mathbb{R}^{h\times w\times c}$ in the samples $z_k$ as
\begin{eqnarray}
    \epsilon_k = \epsilon_\theta(z_k, t, c_p, y),
\end{eqnarray}
where $c_p\in\mathbb{R}^{d}$ is embedding the input text and $y$ is the input segmentation map.
We use the predicted noise $\epsilon_k$ as the input noise to preserve the original information of images. Finally, the initial sample $z_k'\in\mathbb{R}^{h\times w\times c}$ is obtained as
\begin{eqnarray}
    z_k' = \alpha_tz_0+\sigma_t\epsilon_k.
\end{eqnarray}

Then, the noisy sample $z_k'$ is denoised for $k$ steps to generate the images guided by text description and input segmentation map. When generating images, we use a single text description that describes general concepts of target domain images (e.g. driving at night) to guide the generated images to the target domain.

Since our method generates images based on segmentation maps, we can use the original segmentation maps as annotations for the generated images.

\subsubsection*{Model Adaptation:}
Having obtained target-like transferred images, we incorporate them
for training our segmentation model and for learning domain-invariant features. 
To this end, in the training, 
we maximise feature similarity from the feature extractor.

Let $I$ and $I'$ be original and target-like generated images, respectively, and $F$ be a feature extractor for segmentation.
Given feature vectors $F(I)$ and $F(I')\in\mathbb{R}^{D}$, 
we maximize these features in terms of their similarity by
\begin{equation}
    \mathcal{L_{\mathrm{sim}}} = 1 - \frac{F(I)^\top F(I')}{\Vert F(I)\Vert_2\cdot\Vert F(I')\Vert_2}.
\end{equation}

On the other hand,
we use task-specific supervision loss, $\mathcal{L_\mathrm{task}}$, on both the original images and target-like generated images. 
Here, we use the original annotation as ground truth for synthesized images because we generate images based on the annotation. 
Let $H$ be a semantic segmentation decoder, $I$ and $I'$ an image and synthesized image, respectively, and $y$ its semantic label. The task loss can be defined as
\begin{equation}
    \mathcal{L}_\mathrm{task} = \mathrm{CE}(H(F(I)), y) + \mathrm{CE}(H(F(I')), y),
\end{equation}
where $\mathrm{CE}(\cdot, \cdot)$ denotes the categorical cross-entropy loss.

We then calculate the final training objective as:
\begin{equation}
    \mathcal{L} = \lambda\mathcal{L_{\mathrm{sim}}} + \mathcal{L_\mathrm{task}},
\end{equation}
where $\lambda$ is a hyper-parameter\footnote{We set $\lambda = 0.1$ in our experiments.}.

\section{Experiments}
To evaluate the proposed method, we perform semantic segmentation in zero-shot domain adaptation settings 
considering several different scenarios of adaptation: 
{\tt day$\rightarrow$night}, {\tt clear$\rightarrow$snow}, {\tt clear$\rightarrow$rain}, {\tt clear$\rightarrow$fog}, and {\tt real$\rightarrow$game}. 
For the source dataset
we use CityScapes~\cite{Cordts2016Cityscapes}
whereas for the target dataset we use 
ACDC~\cite{sakaridis2021acdc} and GTA5~\cite{richter2016playing} in adverse weather and game domain, respectively.

\subsection{Setup}

For layout-to-image diffusion models, we employ ControlNet-v1.1~\cite{zhang2023adding} trained by ADE-20K~\cite{zhou2017ade20k}, a scene-centric semantic segmentation dataset. For prompts we use the template of ``driving $<$domain$>$''. $<$domain$>$ is a description of the target domain that is selected based on the target domain among: ``at night'', ``in snow'', ``under rain'', ``in fog'', and ``in a game''. We generate a synthetic target-like dataset of cardinality 2975, equivalent to the Cityscapes training set. Each image corresponds one-to-one with the original image. The model generates images at 512$\times$1024 resolutions. These generated images serve as the target-like generated images.

\subsubsection*{Datasets:}
As briefed above we use Cityscapes~\cite{Cordts2016Cityscapes}, composed of 2975 training and 500 validation images, for generating target-like images and training models. For evaluating the robustness of our method against adverse conditions, we use ACDC~\cite{sakaridis2021acdc} which contains urban images captured in adverse conditions: at night, in snow, under rain, or in fog. We also consider a {\tt real$\rightarrow$game} setting; for the dataset of game images, we choose  GTA5~\cite{richter2016playing}, rendered using the open-world video game Grand Theft Auto 5 from the car perspective in the streets. The evaluation is on the validation set for ACDC and a random subset of 1000 images for GTA5.


\subsubsection*{Implementation Details:}
We use the DeepLabv3+~\cite{chen2018encoderdecoder} architecture with the backbone ResNet-50~\cite{He2016-zx} pre-trained by ImageNet-1K~\cite{Deng2009-rk}. The model is trained for 100 epochs on random 384$\times$768 crops with batch size 4. We use a polynomial learning rate schedule with an initial learning rate $10^{-3}$. 
SGD with momentum 0.9 and weight decay $10^{-4}$
for optimization. 
As for data augmentation we apply standard color jittering and horizontal flips to crops. For evaluation, we use the final checkpoint in the last epoch. 
All the experiments have been conducted on NVIDIA V100.

\subsection{Main Results}
\label{sec: main results}
For all the experiments of segmentation, we report the mean Intersection over Union (mIoU) results by taking the mean of three independent trials on different random seeds.
We discuss the results by our method, ZoDi, while comparing them  with two state-of-the-art baselines: 
P\O DA~\cite{fahes2023poda} and DATUM~\cite{benigmim2023datum} (P\O DA is a zero-shot domain adaptation method with CLIP and DATUM is a one-shot with diffusion models, see Sec.~\ref{related} for more details). 
We also make a comparison to DAFormer~\cite{hoyer2022daformer} for a strong baseline of unsupervised domain adaptation where 
full images from the target domain are accessible.
This work uses simple prompts to describe the target domains, e.g. ``driving $<$domain$>$''. 
We also conduct source-only training with the same conditions for comparing ZoDi with the models trained only with source images.
The results are summarised in Tab.~\ref{table:poda}.

\begin{table*}[t]
\begin{center}
\caption{\textbf{The results of segmentation} by different methods for five adaptation scenarios (in mIoU). $^\dag$Our reimplementation.}
\scalebox{0.9}{
\label{table:poda}
\begin{tabular}{@{}lcclllll@{}} \toprule
     & \#Target & Model & {\tt Day} & {\tt Clear} & {\tt Clear} & {\tt Clear} & {\tt Real} \\
    Method & Images & Agnostic & $\rightarrow${\tt Night} & $\rightarrow${\tt Snow} & $\rightarrow${\tt Rain} & $\rightarrow${\tt Fog} & $\rightarrow${\tt Game} \\ \midrule
    Source-only & - & - & 20.7\fontsize{7}{7}$\pm1.0$ & 40.8\fontsize{7}{7}$\pm0.3$ & 39.4\fontsize{7}{7}$\pm0.3$ & 54.5\fontsize{7}{7}$\pm1.0$ & 40.1\fontsize{7}{7}$\pm0.7$\\
    DAFormer~\cite{hoyer2022daformer} & 400 & \cmark & 30.3$^\dag$\fontsize{7}{7}$\pm0.3$ & 44.5$^\dag$\fontsize{7}{7}$\pm1.3$ & 41.0$^\dag$\fontsize{7}{7}$\pm2.2$ & 58.2$^\dag$\fontsize{7}{7}$\pm0.3$ & 47.9$^\dag$\fontsize{7}{7}$\pm0.6$ \\ \cmidrule{1-8}
    DATUM+DAFormer~\cite{benigmim2023datum} & one & \cmark & 19.9$^\dag$\fontsize{7}{7}$\pm0.6$ & 43.0$^\dag$\fontsize{7}{7}$\pm0.1$ & 39.0$^\dag$\fontsize{7}{7}$\pm1.5$ & 60.3$^\dag$\fontsize{7}{7}$\pm0.4$ & 44.7$^\dag$\fontsize{7}{7}$\pm0.7$ \\ \cmidrule{1-8}
    P\O DA~\cite{fahes2023poda} & zero & \xmark & \textbf{25.0}\fontsize{7}{7}$\pm0.5$ & 43.9\fontsize{7}{7}$\pm0.5$ & 42.3\fontsize{7}{7}$\pm0.6$ & 49.0$^\dag$\fontsize{7}{7}$\pm0.9$ & \textbf{41.1}\fontsize{7}{7}$\pm0.5$ \\
    ZoDi (Ours) & zero & \cmark & 24.7\fontsize{7}{7}$\pm0.2$ & \textbf{45.6}\fontsize{7}{7}$\pm1.6$ & \textbf{47.0}\fontsize{7}{7}$\pm1.2$ & \textbf{56.1}\fontsize{7}{7}$\pm1.7$ & 40.5\fontsize{7}{7}$\pm0.2$ \\ \bottomrule
    
\end{tabular}
}
\end{center}
\end{table*}

\subsubsection*{Comparison to Zero-Shot Method (P\O DA):}
We use the same architecture, prompts, data augmentation, and optimizer as in P\O DA for fair comparisons. 
ZoDi outperforms P\O DA in three settings by large margins: 45.6 vs. 43.9 in {\tt clear$\rightarrow$snow}, 47.0 vs. 42.3 in {\tt clear$\rightarrow$rain}, and 56.1 vs. 49.0 in {\tt clear$\rightarrow$fog}. In the other two settings, {\tt day$\rightarrow$night} and {\tt real$\rightarrow$game}, P\O DA shows a slight edge over ZoDi (24.7 vs. 25.0 and 40.5 vs. 41.1, respectively). However, it is important to note that the performance gap in these scenarios is relatively small. Despite P\O DA's marginally better results in these cases, ZoDi still surpasses models trained solely on source data (24.7 vs. 20.7 and 40.5 vs. 40.1, respectively). This trend holds consistently across all settings, highlighting ZoDi's robust performance even in cases where it is slightly outperformed by P\O DA.

\begin{figure*}[t]
\begin{center}
\hspace*{-10mm}
   \includegraphics[width=1.0\linewidth]{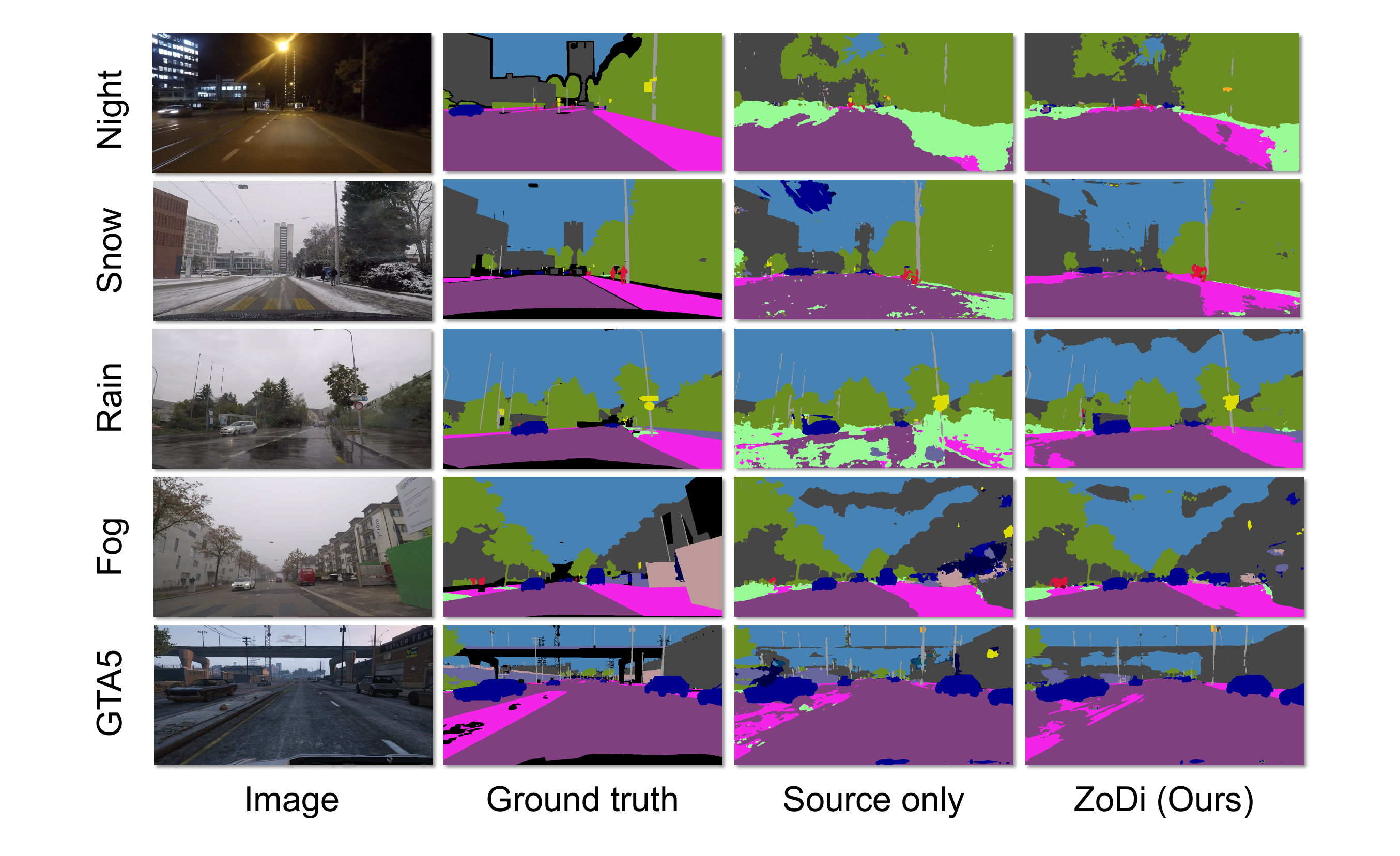}
   \caption{\textbf{The qualitative results by different methods.} The results on each dataset (ACDC or GTA5) are shown. Compared with models trained only by source images (column 3), ZoDi (rightmost) delivers segmentation closer to the ground truth.} 
   \label{fig:qualitative_results}
\end{center}
\end{figure*}

When it comes to the settings where ZoDi does not outperform P\O DA ({\tt day$\rightarrow$night} and {\tt real$\rightarrow$game}), it could be attributed to the difficulties in drastically changing the domains. As can be seen in Fig.~\ref{fig:generated_images}, it can be challenging for ZoDi to make severe changes. For example, in {\tt day$\rightarrow$night} setting, ZoDi appears to find it hard to darken images to a large extent while keeping the image content. In {\tt real$\rightarrow$game} setting, it struggles with the discrepancy between target images (as seen in Fig.~\ref{fig:qualitative_results}) and generated images. 

However, it is worth noting that ZoDi is more available than P\O DA in the following points. While P\O DA requires freezing the backbone of CLIP, ZoDi does not assume any specific models and hence can be used for existing daytime models. In addition, we can visualize and analyze the generated images. Unlike P\O DA, which adapts the model in the CLIP feature space, ZoDi generates an actual image. This means that the user can explore suitable prompts or other image transfer methods for a better performance and estimate the model's performance without target images. 
These are beneficial features of ZoDi over existing CLIP-based domain adaptation methods.

\subsubsection{Qualitative Results of Zero-Shot Image Transfer:}
\begin{figure*}[t]
\begin{center}
   \includegraphics[width=1.05\linewidth]{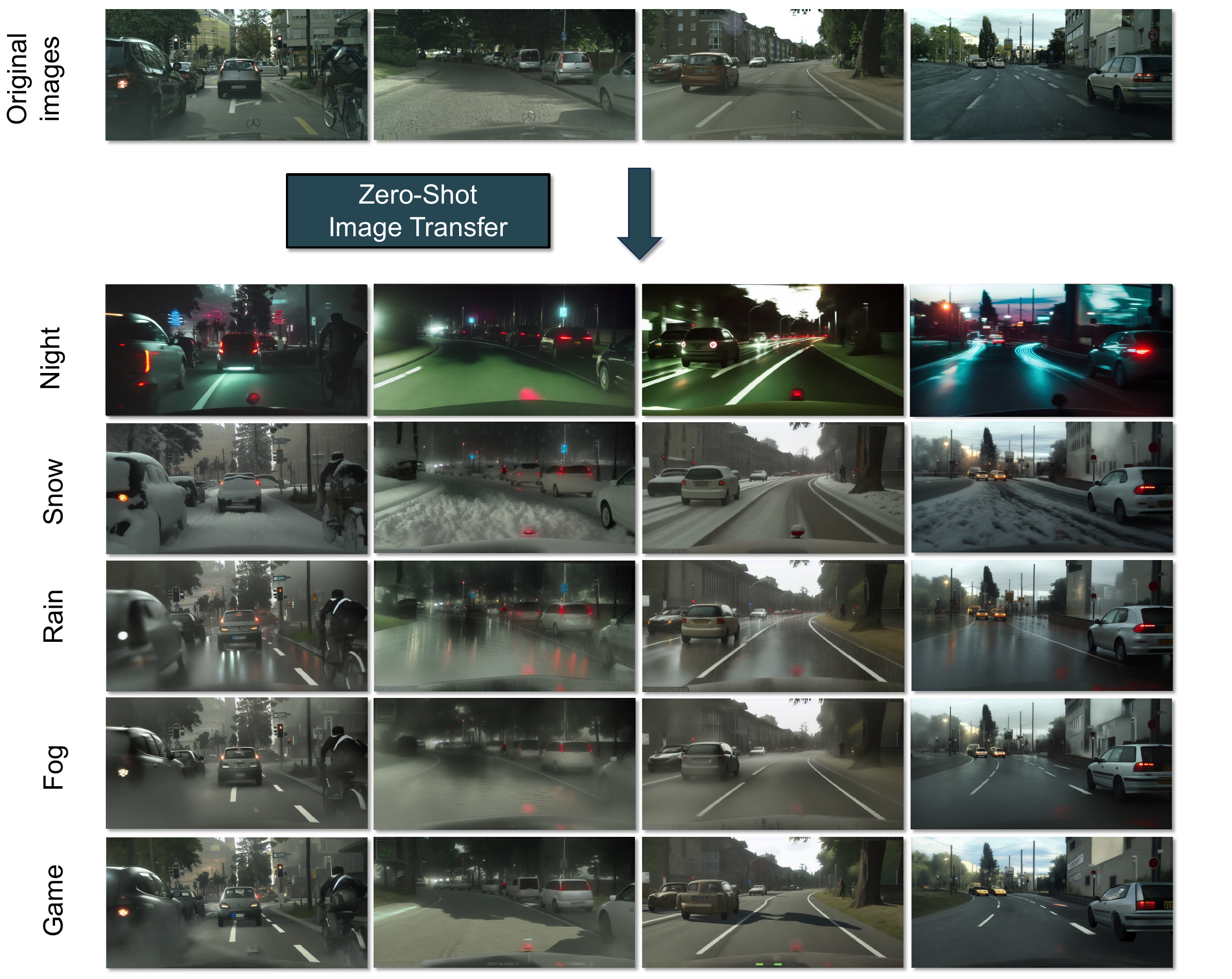}
\end{center}
   \caption{\textbf{Examples of generated images.} Top: four original images from CityScapes. Bottom: generated images for each of the four by our zero-shot image transfer into five different domains.
   }
\label{fig:generated_images}
\end{figure*}
Fig.~\ref{fig:generated_images} shows some of target-like images generated by our zero-shot image transfer which makes a conversion given the original images in the top row. Note that our method successfully generates high-quality target-like images while keeping their layout and content. For example, the roads and cars are covered with snow in generated images for the snow domain (the third row) whereas the original images (top row) do not contain snow. In addition, in the generated images we do not lose the quality or objects in the original images. These results exemplify the quality by our zero-shot image transfer method.

\subsubsection*{Comparison to UDA Method (DATUM with DAFormer):}
We train DATUM with one random image from the target domain. As shown in Tab.~\ref{table:poda}, ZoDi outperforms DATUM with DAFormer in three settings (24.7 vs. 19.9 in {\tt day$\rightarrow$night}, 45.8 vs. 43.0 in {\tt clear$\rightarrow$snow}, and 47.0 vs. 39.0 in {\tt clear$\rightarrow$rain}), while it is the contrary in other two settings (56.1 vs. 60.3 in {\tt clear$\rightarrow$fog} and 40.5 vs. 44.7 in {\tt real$\rightarrow$game}). 
Notice that DATUM harms the original performance (20.7 to 19.9 in {\tt day$\rightarrow$night} and 39.4 to 39.0 in {\tt clear$\rightarrow$rain}).
Here, it is also worth noting that our ZoDi does not require any images from the target domain, nor does it rely on other complicated, unsupervised domain adaptation methods.

We also compare ZoDi with DAFormer for a strong baseline of unsupervised domain adaptation, where full images from the target domain are accessible. Since the number of training images in ACDC is 400 in each domain, we randomly select 400 images for training in GTA5 for a fair comparison.
We train DAFormer using 400 target images without labels for 20k iterations. This number of iterations is smaller than the original paper (i.e. 40k) because we found that training for longer iterations leads to overfitting due to the size of the dataset. Tab.~\ref{table:poda} shows DAFormer can strongly improve the performance on the target domain compared with the models trained only by source images (e.g. 20.7 to 30.0 in {\tt day$\rightarrow$night} and 54.5 to 57.8 in {\tt clear$\rightarrow$fog}). 
Surprisingly, ZoDi surpasses DAFormer in some cases (45.6 vs. 43.8 in {\tt clear$\rightarrow$snow}, 47.0 vs. 42.5 in {\tt clear$\rightarrow$rain}) although it does not require any target images. 

\begin{figure*}[h]
\begin{center}
   \includegraphics[width=1.0\linewidth]{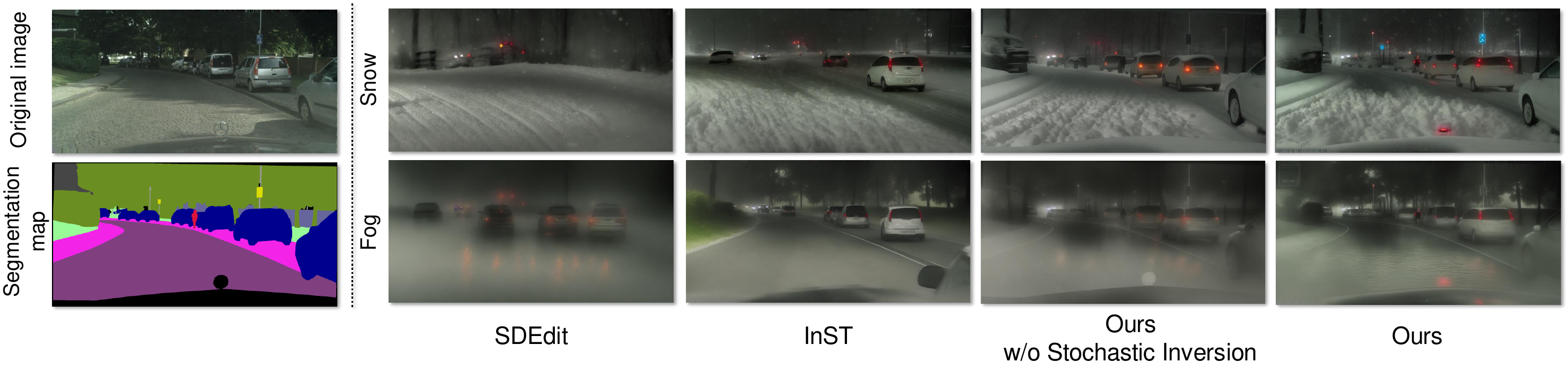}
   \caption{\textbf{Ablation studies for our zero-shot image transfer method.} Our method consisting of ControlNet and Stochastic Inversion is highly capable of changing the domain of the images as can be seen in the rightmost column. In contrast, the images generated without either ControlNet or Stochastic Inversion (column 2, column 3, or column 4) can collapse the original contents.}
   \label{fig:ablation_image_transfer}
\end{center}
\end{figure*}

\subsubsection*{Qualitative Results of ZoDi:}

Fig.~\ref{fig:qualitative_results} shows qualitative segmentation results on each dataset. 
ZoDi extracts information from various domains and thus can be considered to generate more accurate segmentation maps. For example, despite the rainy condition (row\#3) it recognizes the road more correctly while the model trained only by source images obviously fails to capture the information. 

However, we also observe some rooms for improvement, e.g. ZoDi still struggles in recognizing the sky in the {\tt day$\rightarrow$night} setting (row\#1). This can be attributed to the brighter sky in the generated images (Fig.~\ref{fig:generated_images}) compared with the original night images (Fig.~\ref{fig:qualitative_results}). As mentioned earlier, changing the domain drastically can be challenging. Exploring more robust image transfer methods could lead to further improvements in performance.

\subsection{Ablation Studies}

\subsubsection*{Zero-Shot Image Transfer:}
We compare our diffusion-based image transfer method with SDEdit~\cite{meng2022sdedit}, InST~\cite{zhang2023inversionbased}, and ours without SI (Stochastic Inversion). SDEdit is equivalent to the method of our image transfer but without ControlNet and stochastic inversion, and InST is likewise without ControlNet. Fig.~\ref{fig:ablation_image_transfer} shows the generated images by four methods including ours. Among those the qualitative results illustrate that our transfer method best maintains the original layout and contents. In our quantitative evaluation, as shown in Tab.~\ref{table:ablation}, we replace our method with SDEdit, InST, or ours without SI while keeping the model-level adaptation stage unchanged. Our image transfer method in ZoDi outperforms the other alternatives. We also compare with other image transfer methods (Prompt-to-Prompt~\cite{hertz2022prompttoprompt} and InstructPix2Pix) in supplementary.

\subsubsection*{Model Adaptation:}
Turning to our model adaption architecture, 
we examine effect of similarity loss, $L_\mathrm{sim}$, on the performance by opting it out (see Tab.~\ref{table:ablation}).
We see that the performance 
still surpasses the model trained only by the source original images. 
This implies that our transfer method well serves as a method of data augmentation. Finally, we find that combining model adaptation boosts performance in most cases except for {\tt clear$\rightarrow$fog} setting. 
These results 
support the effectiveness of our model adaptation architecture.



\subsubsection*{Different Architectures:}
We applied our method to other backbones (ResNet-101) and decoders for segmentation (RefineNet~\cite{lin2016refinenet})
for studying possible dependency on different architectures.
Tab.~\ref{table:different_architecture} shows the results for all the five settings where our method brings consistent improvements, indicating its robustness across different architectures.

\begin{table*}[t]
\vspace{5mm}
\begin{center}
\caption{\textbf{Ablation studies for each component.}}
\scalebox{0.9}{
\label{table:ablation}
\begin{tabular}{@{}lllllll@{}} \toprule
     & & {\tt Day} & {\tt Clear} & {\tt Clear} & {\tt Clear} & {\tt Real} \\
    \small{Category} & \small{Method} & {\tt $\rightarrow$Night} & {\tt $\rightarrow$Snow} & {\tt $\rightarrow$Rain} & {\tt $\rightarrow$Fog} & {\tt $\rightarrow$Game} \\ \midrule
    Source-only & DeepLabv3+ & 20.7\fontsize{7}{7}$\pm1.0$ & 40.8\fontsize{7}{7}$\pm0.3$ & 39.4\fontsize{7}{7}$\pm0.3$ & 54.5\fontsize{7}{7}$\pm1.0$ & 40.1\fontsize{7}{7}$\pm0.7$ \\ \cmidrule{1-7}
    \multirow{3}{*}{Image transfer} & SDEdit~\cite{meng2022sdedit} & 18.2\fontsize{7}{7}$\pm0.7$ & 36.0\fontsize{7}{7}$\pm3.2$ & 35.4\fontsize{7}{7}$\pm0.9$ & 46.4\fontsize{7}{7}$\pm1.3$ & 37.2\fontsize{7}{7}$\pm0.1$\\
     & InST~\cite{zhang2023inversionbased} & 16.3\fontsize{7}{7}$\pm0.9$ & 34.8\fontsize{7}{7}$\pm2.5$ & 37.6\fontsize{7}{7}$\pm2.0$ & 51.1\fontsize{7}{7}$\pm2.0$ & 37.6\fontsize{7}{7}$\pm0.7$\\
 & w/o SI~\cite{zhang2023inversionbased} & 24.3\fontsize{7}{7}$\pm0.9$ & 43.1\fontsize{7}{7}$\pm0.9$ & 45.0\fontsize{7}{7}$\pm1.0$ & 53.9\fontsize{7}{7}$\pm1.1$ & 39.5\fontsize{7}{7}$\pm0.6$ \\
    Similarity loss & w/o $L_\mathrm{sim}$ & 24.5\fontsize{7}{7}$\pm0.5$ & 44.0\fontsize{7}{7}$\pm1.8$ & 46.0\fontsize{7}{7}$\pm0.7$ & \textbf{56.4}\fontsize{7}{7}$\pm2.0$ & 40.4\fontsize{7}{7}$\pm0.2$ \\ \cmidrule{1-7}
    Full-version & ZoDi (Ours) & \textbf{24.7}\fontsize{7}{7}$\pm0.2$ & \textbf{45.6}\fontsize{7}{7}$\pm1.6$ & \textbf{47.0}\fontsize{7}{7}$\pm1.2$ & 56.1\fontsize{7}{7}$\pm1.7$ & \textbf{40.5}\fontsize{7}{7}$\pm0.2$ \\ \bottomrule
    
\end{tabular}
}
\end{center}
\end{table*}

\begin{table*}[t]
\begin{center}
\caption{\textbf{Different architectures.}}
\scalebox{0.9}{
\label{table:different_architecture}
\begin{tabular}{@{}llllllll@{}} \toprule
     & & & {\tt Day} & {\tt Clear} & {\tt Clear} & {\tt Clear} & {\tt Real} \\
    \small{Model} & \small{Backbone} & \small{Method} & {\tt $\rightarrow$Night} & {\tt $\rightarrow$Snow} & {\tt $\rightarrow$Rain} & {\tt $\rightarrow$Fog} & {\tt $\rightarrow$Game} \\ \midrule
    \multirow{2}{*}{RefineNet}& \multirow{2}{*}{ResNet-50} & source-only & 24.4\fontsize{7}{7}$\pm0.4$ & 46.0\fontsize{7}{7}$\pm1.2$ & 42.3\fontsize{7}{7}$\pm0.9$ & 60.6\fontsize{7}{7}$\pm0.4$ & 44.1\fontsize{7}{7}$\pm0.7$\\
    & & ZoDi (Ours) & \textbf{27.1}\fontsize{7}{7}$\pm0.5$ & \textbf{46.4}\fontsize{7}{7}$\pm0.2$ & \textbf{46.6}\fontsize{7}{7}$\pm0.5$ & \textbf{62.4}\fontsize{7}{7}$\pm0.5$ & \textbf{44.2}\fontsize{7}{7}$\pm0.6$\\ \cmidrule{1-8}
    \multirow{4}{*}{DeepLabv3+}& \multirow{2}{*}{ResNet-50} & source-only & 20.7\fontsize{7}{7}$\pm1.0$ & 40.8\fontsize{7}{7}$\pm0.3$ & 39.4\fontsize{7}{7}$\pm0.3$ & 54.5\fontsize{7}{7}$\pm1.0$ & 40.1\fontsize{7}{7}$\pm0.7$ \\
    & & ZoDi (Ours) & \textbf{24.7}\fontsize{7}{7}$\pm0.2$ & \textbf{45.6}\fontsize{7}{7}$\pm1.6$ & \textbf{47.0}\fontsize{7}{7}$\pm1.2$ & \textbf{56.1}\fontsize{7}{7}$\pm1.7$ & \textbf{40.5}\fontsize{7}{7}$\pm0.2$ \\ \cmidrule{2-8}
    & \multirow{2}{*}[-2pt]{ResNet-101} & source-only & 23.0\fontsize{7}{7}$\pm0.5$ & 44.4\fontsize{7}{7}$\pm0.4$ & 42.2\fontsize{7}{7}$\pm0.6$ & 57.9\fontsize{7}{7}$\pm0.4$ & 41.1\fontsize{7}{7}$\pm0.2$ \\
    & & ZoDi (Ours) & \textbf{25.6}\fontsize{7}{7}$\pm0.7$ & \textbf{45.8}\fontsize{7}{7}$\pm0.4$ & \textbf{45.4}\fontsize{7}{7}$\pm0.2$ & \textbf{59.2}\fontsize{7}{7}$\pm1.4$ & \textbf{41.5}\fontsize{7}{7}$\pm0.5$\\ \bottomrule
    
\end{tabular}
}
\end{center}
\end{table*}

\subsection{Failure Cases}
\begin{figure*}[t]
\begin{center}
\vspace{5mm}
   \includegraphics[width=1.04\linewidth]{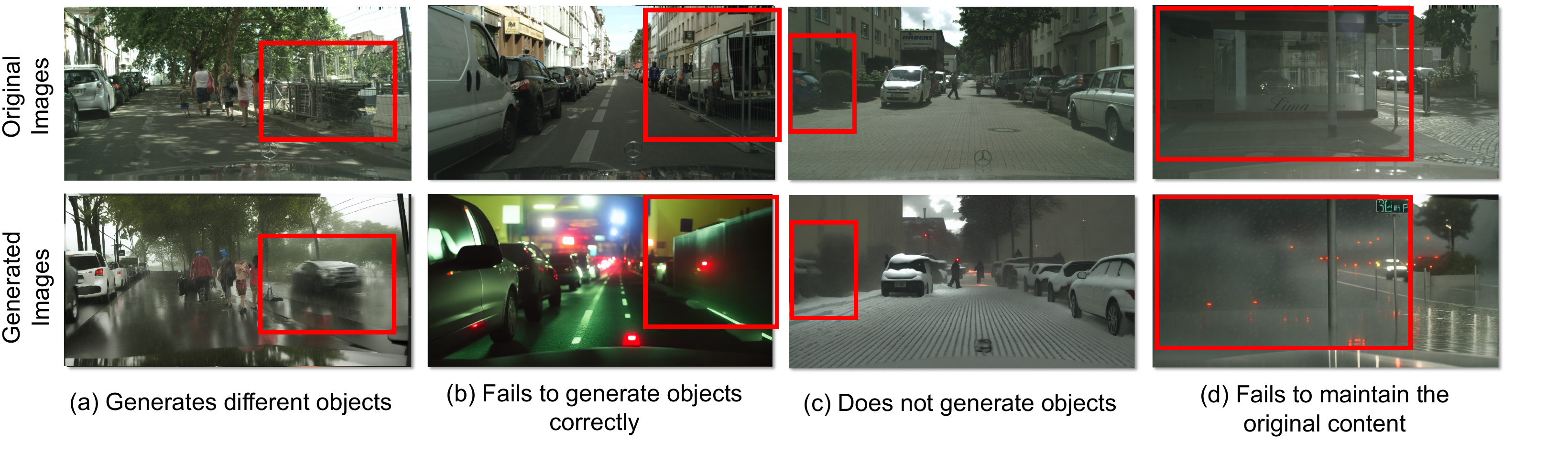}
\end{center}
   \caption{\textbf{Failure cases in our zero-shot image transfer.} (a) Generates different objects (car) in {\tt clear$\rightarrow$rain}. (b) Fails to generate objects correctly (trucks) in {\tt day$\rightarrow$night}. (c) Does not generate objects (car) in {\tt day$\rightarrow$snow}. (d) Fails to maintain the original content in {\tt day$\rightarrow$fog}.}
\label{fig:failure_cases}
\end{figure*}
We study some failure cases of image transfer in ZoDi because the quality of generated images is of importance. As in Fig.~\ref{fig:failure_cases}, our method sometimes make errors such that it could generate other different objects, fail to generate objects correctly, not generate objects, or fail to maintain the original content of the images. In addition, in the {\tt day$\rightarrow$night} setting as in Fig.~\ref{fig:failure_cases}(b), it tends to fail to keep the original image content.

It can however be seen as an advantage that our image transfer allows us to visualize and explore the reason for the models' performance when it does not perform perfectly. This visualization is not available for CLIP-based domain adaptation methods since they conduct domain adaptation in the CLIP feature space. 

\section{Conclusion}
This paper introduced zero-shot domain adaptation by diffusion-based image transfer, called ZoDi, while addressing the critical domain-shift problem in segmentation tasks. ZoDi leverages powerful diffusion models to transfer source images to the target domain in a zero-shot manner. Its components, namely image transfer and model adaptation, work synergistically to create domain-robust representations for a segmentation model.

The experiments demonstrate that the performance of ZoDi exceeds that of existing zero-shot method. In particular, the utilization of layout-to-image diffusion models, guided by real images and complemented by stochastic inversion technique, leads to the successful performance; it outperforms current state of the art on average while providing a more flexible and powerful solution to some challenges in zero-shot domain adaptation.

Although the proposed image transfer in ZoDi allows us to generate high-quality images, it can also fail, e.g. to correctly generate specific objects. As suggested in Sec.~\ref{sec: main results}, some drastic changes of domain are yet beyond its capacity, leaving a need for further accurate image transfer in future development.
%

In summary, nevertheless, we believe that the paper contributes to extend the availability of zero-shot domain adaptation by presenting ZoDi as a promising approach with practical implications for when obtaining target images can be challenging in real-world applications.
We hope this research helps open new avenues for enhancing adaptability of deep learning models by leveraging synthetic data generated by diffusion models. 

\clearpage
%
%
\bibliographystyle{splncs04}
\bibliography{egbib}

\clearpage
\newcommand\beginsupplement{%
        \setcounter{table}{0}
        \renewcommand{\thetable}{\Alph{table}}%
        \setcounter{figure}{0}
        \renewcommand{\thefigure}{\Alph{figure}}%
     }
\beginsupplement
\appendix

\begin{figure*}[h]
\begin{center}
   \includegraphics[width=1.00\linewidth]{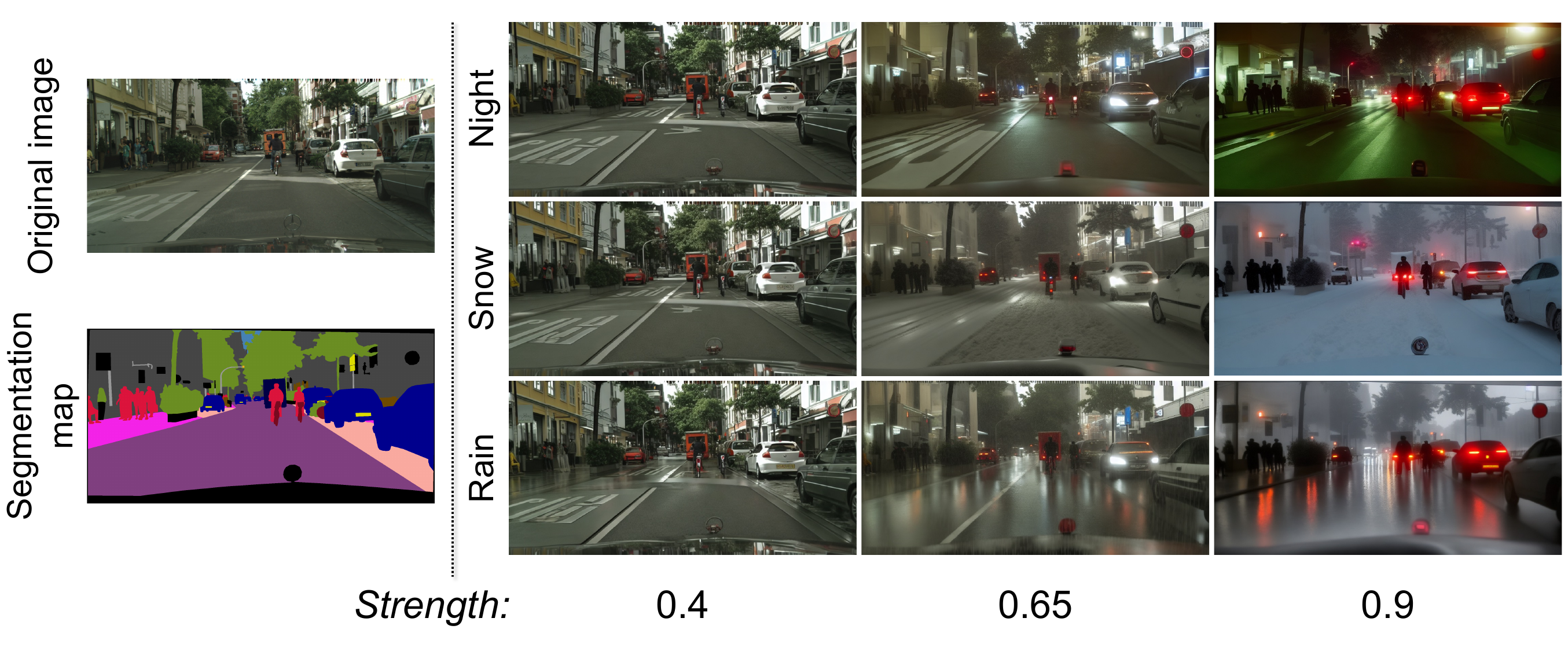}
   \caption{\textbf{Generated images based on different \textit{strength}.} We show  images in night, snow, and rain domain generated based on different \textit{strength}: 0.4, 0.65, and 0.9.
    }
   \label{fig:strength}
\end{center}
\end{figure*}
\section{Hyper-parameter \textit{Strength}}
The most relevant hyper-parameter for image synthesis is the \textit{strength} of changes. The \textit{strength} is directly related to the value of $k$ defined in Sec.3.2, which is used for the number of steps in the diffusion process. Let $T\in\mathcal{N}$ be the number of whole time steps and $S\in [0, 1]$ be \textit{strength}. $k$ is then calculated as
\begin{equation}
    k = \lfloor TS\rfloor,
\end{equation}
where $\lfloor \cdot \rfloor$ denotes the floor function. The greater the value of \textit{strength}, the stronger the influence of the target text prompt on the generated images, and vice versa; the generated images remain closer to the original image.

The generated images based on different \textit{strength}s are shown in Fig.~\ref{fig:strength}. 
The figure illustrates that a greater value of \textit{strength} makes bigger changes although it could result in losing some original content. We set the \textit{strength} as $0.6$ for the fog and game domain, $0.65$ for the snow, and rain domain, and $0.9$ for the night domain.
We do not use images from the target domain to explore the hyper-parameters because 
it is possible 
by inspecting generated images.

\section{Comparison with other image transfer methods}

\begin{figure*}[h]
\begin{center}
   \includegraphics[width=1.00\linewidth]{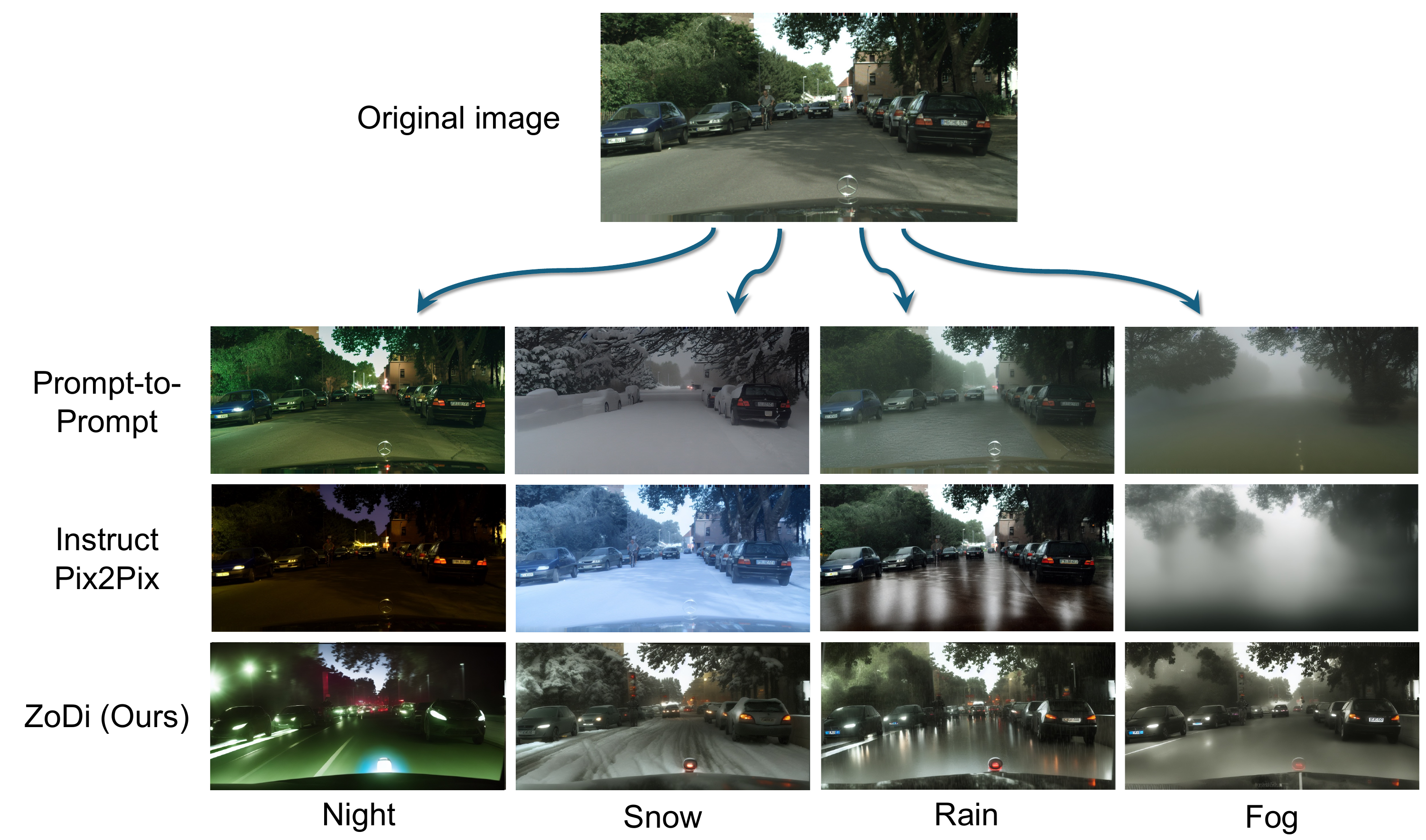}
   \caption{\textbf{Generated images by ZoDi and other image transfer methods (Prompt-to-Prompt~\cite{hertz2022prompttoprompt} and InstructP2P~\cite{Brooks2023instructpix2pix}}) We show images in night, snow, rain and fog domain.
    }
   \label{img:comparison}
\end{center}
\end{figure*}

\begin{table}[t]
\begin{center}
\small
\caption{Comparison with other image transfer methods}
\label{table:comparison}
\begin{tabular}{@{}llll@{}} \toprule
     & & CLIP & \\
    Domain & Method & Similarity & mIoU (\%) \\ \midrule
    \multirow{4}{*}[-2pt]{\tt Day$\rightarrow$Night} & Source-only & 20.5\fontsize{7}{7}$\pm1.6$ & 8.2\fontsize{7}{7}$\pm0.1$ \\
    & Prompt-to-Prompt~\cite{hertz2022prompttoprompt} & 27.4\fontsize{7}{7}$\pm2.1$ & \textbf{9.5}\fontsize{7}{7}$\pm0.3$\\
    & InstructPix2Pix~\cite{Brooks2023instructpix2pix} & 28.4\fontsize{7}{7}$\pm1.8$ & 8.2\fontsize{7}{7}$\pm0.5$ \\
    & ZoDi (Ours) & \textbf{30.0}\fontsize{7}{7}$\pm1.1$ & 9.1\fontsize{7}{7}$\pm0.5$ \\ \cmidrule{1-4}
    \multirow{4}{*}[-2pt]{\tt Clear$\rightarrow$Snow} & Source-only & 20.2\fontsize{7}{7}$\pm1.3$ & 14.7\fontsize{7}{7}$\pm1.4$ \\ 
    & Prompt-to-Prompt~\cite{hertz2022prompttoprompt} & 26.1\fontsize{7}{7}$\pm1.5$ & 16.5\fontsize{7}{7}$\pm0.3$ \\
    & InstructPix2Pix~\cite{Brooks2023instructpix2pix} & 25.9\fontsize{7}{7}$\pm2.1$ & 16.6\fontsize{7}{7}$\pm0.2$ \\
    & ZoDi (Ours) & \textbf{28.2}\fontsize{7}{7}$\pm1.8$ & \textbf{18.1}\fontsize{7}{7}$\pm0.3$ \\ \cmidrule{1-4}
    \multirow{4}{*}[-2pt]{\tt Clear$\rightarrow$Rain} & Source-only & 22.0\fontsize{7}{7}$\pm1.4$ & 16.6\fontsize{7}{7}$\pm0.3$ \\ 
    & Prompt-to-Prompt~\cite{hertz2022prompttoprompt} & 27.7\fontsize{7}{7}$\pm1.4$ & 18.2\fontsize{7}{7}$\pm0.2$ \\
    & InstructPix2Pix~\cite{Brooks2023instructpix2pix} & 28.5\fontsize{7}{7}$\pm1.4$ & \textbf{19.2}\fontsize{7}{7}$\pm0.2$ \\
    & ZoDi (Ours) & \textbf{29.7}\fontsize{7}{7}$\pm1.1$ & 18.9\fontsize{7}{7}$\pm0.1$ \\ \cmidrule{1-4}
    \multirow{4}{*}[-2pt]{\tt Clear$\rightarrow$Fog} & Source-only & 22.0\fontsize{7}{7}$\pm1.4$ & 16.4\fontsize{7}{7}$\pm0.7$ \\ 
    & Prompt-to-Prompt~\cite{hertz2022prompttoprompt} & \textbf{30.5}\fontsize{7}{7}$\pm2.2$ & 15.7\fontsize{7}{7}$\pm0.8$ \\
    & InstructPix2Pix~\cite{Brooks2023instructpix2pix} & 30.1\fontsize{7}{7}$\pm1.8$ & 16.1\fontsize{7}{7}$\pm0.4$ \\
    & ZoDi (Ours) & 28.8\fontsize{7}{7}$\pm2.7$ & \textbf{19.4}\fontsize{7}{7}$\pm0.5$ \\ \bottomrule
\end{tabular}
\end{center}
\end{table}

We also compare our image transfer methods with Prompt-to-Prompt~\cite{hertz2022prompttoprompt} and InstructPix2Pix~\cite{Brooks2023instructpix2pix}.
Our approach differs from others by focusing on image generation to train segmentation models. Unlike other methods, our approach generates images based on segmentation maps. Therefore, our method is more accurate against the segmentation maps (Tab.~\ref{img:comparison}). Especially, the layout of images collapses in the fog domain in other methods, while ZoDi succeeds in keeping the layout.

Given that Prompt-to-Prompt and InstructPix2Pix each take about 3 minutes to generate an image on an NVIDIA A100, generating all images corresponding to Cityscapes~\cite{Cordts2016Cityscapes} would be prohibitively time-consuming. Therefore, we generate 100 images for each domain and each image transfer method. In Prompt-to-Prompt, we use ``driving'' as the source prompt and ``driving $<$domain$>$'' as the target prompt, the same as the prompt we used in Sec. 4.1. $<$domain$>$ is a description of the target domain. In InstructPix2Pix, we use ``make it $<$domain$>$'' as the instruction prompt.

First, we evaluate the image transfer methods using CLIP similarity~\cite{Radford2021-bi}, which measures the similarity between the image and text prompt. The table~\ref{table:comparison} shows that our method best aligns with the text prompt on average. Second, we reported the mIoU score achieved by models trained using the original 100 images and the generated 100 images. ZoDi shows consistent improvement and outperforms other methods on average.

Our method not only excels in aligning generated images with textual prompts but also enhances the performance of segmentation models. This demonstrates our approach's effectiveness in training segmentation models in domain adaptation settings.

\end{document}